\documentclass{article}

\usepackage{arxiv}

\usepackage[utf8]{inputenc} % allow utf-8 input
\usepackage[T1]{fontenc}    % use 8-bit T1 fonts
\usepackage{hyperref}       % hyperlinks
\usepackage{url}            % simple URL typesetting
\usepackage{booktabs}       % professional-quality tables
\usepackage{amsfonts}       % blackboard math symbols
\usepackage{nicefrac}       % compact symbols for 1/2, etc.
\usepackage{microtype}      % microtypography
\usepackage{lipsum}
\usepackage{graphicx}
\usepackage{comment}
\usepackage{float}
\usepackage{rotating}
\graphicspath{ {./images/} }
\usepackage{graphicx}
\usepackage{amsmath}
\usepackage{amssymb}
\usepackage{booktabs}
\usepackage{amsmath,amsfonts,bm}

\def\eqref#1{equation~\ref{#1}}
\def\1{\bm{1}}
\newcommand{\train}{\mathcal{D}}

\DeclareMathAlphabet{\mathsfit}{\encodingdefault}{\sfdefault}{m}{sl}
\SetMathAlphabet{\mathsfit}{bold}{\encodingdefault}{\sfdefault}{bx}{n}

\def\sA{{\mathbb{A}}}
\def\sB{{\mathbb{B}}}
\def\sC{{\mathbb{C}}}

\def\sN{{\mathbb{N}}}

\usepackage{algorithm}
\usepackage{algpseudocode}
\usepackage{hyperref}
\usepackage{tikz}
\usepackage{url}
\usepackage{xcolor}
\usepackage{latexsym}
\usepackage{multicol,multirow}
\usepackage{amsmath,amssymb,amsfonts}
\usepackage{mathrsfs}
\usepackage{amsthm}
\usepackage{rotating}
\usepackage{float}

\title{LightDepth: A Resource Efficient Depth Estimation Approach for Dealing with Ground Truth Sparsity via Curriculum Learning}

\author{
 Fatemeh Karimi \\
  Department of Mathematical Sciences\\
  Sharif University of Technology\\
  Tehran, Iran \\
  \texttt{fatemeh.karimii@sharif.edu} \\
   \And
 Amir Mehrpanah\\
    Department of Mathematical Sciences\\
    Shahid Beheshti University\\
    Tehran, Iran \\
    \texttt{a.mehrpanah@mail.sbu.ac.ir} \\
  \And
 Reza Rawassizadeh  \\
  Department of Computer Science\\
  Boston University\\
  Massachusetts, United States \\
  \texttt{rezar@bu.edu } \\
}

\begin{document}
\maketitle

\begin{abstract}
Advances in neural networks enable tackling complex computer vision tasks such as depth estimation of outdoor scenes at unprecedented accuracy. Promising research has been done on depth estimation. However, current efforts are computationally resource-intensive and do not consider the resource constraints of autonomous devices, such as robots and drones. In this work, we present a “fast” and “battery-efficient” approach for depth estimation. Our approach devises model-agnostic curriculum-based learning for depth estimation. Our experiments show that the accuracy of our model performs on par with the state-of-the-art models, while its response time outperforms other models by 71\%. All codes are available online at \url{https://github.com/fatemehkarimii/LightDepth}.
\end{abstract}

%%%%%%%%% BODY TEXT
\section{Introduction}
\label{sec:introduction}
In early 2012, due to the introduction of the deep neural network, we observed a revolution in machine learning algorithms and some of their applications, including computer vision. Convolutional Neural Networks (CNNs) and recently vision transformer (ViTs)~\cite{han2022survey}, have been used as a de-facto standard in many computer vision tasks~\cite{khan2018guide}. Depth estimation is one of the most challenging tasks in computer vision~\cite{zhan2018unsupervised}. It has many real-world applications, including self-driving cars~\cite{cui2019real}, 3D reconstruction~\cite{diamantas2010depth}, robotic grasping~\cite{walz2020uncertainty}, and scene understanding~\cite{chen2019towards}. 
Despite all advances, ground truth sparsity is one of the problems in computer vision that causes a significant performance loss~\cite{uhrig2017sparsity}. However, sparsity is inevitable~\cite{geiger2013vision} in some applications such as depth estimation, where depth maps are recorded by discrete polar scanning technologies such as Light Detection and Ranging (LiDAR) laser scans ~\cite{lefsky2002lidar} The LiDAR technology produces sparse ground truth depth maps, which makes the model training complicated and error-prone. This problem has been overlooked in recent literature~\cite{alhashim2018high,dorn,adabins, bts}, and researchers have been trying to improve their models' performance by increasing the number of parameters.
As the number of parameters increases, the model becomes resource-intensive.  High-performance computer systems have no major problems with heavy models ~\cite{fedak2001xtremweb, schroeder2009large}. However, these models can not be used for small devices  with constrained resources like wearables, mobile phones and Internet of Things devices ~\cite{gan2021, rong2022}. The need for high computation capability pushes existing depth estimation algorithms to heavy machines only. 

Resources and run-time are essential parameters for applications that operate in real-time. For example, in self-driving cars, the inefficiency of these models or the delay in recognizing the depth may cause a calamitous event.

To address this limitation, we propose a novel \emph{resource-efficient} depth estimation algorithm based on \emph{curriculum learning}. The curriculum is a training policy that trains the model progressively, starting from less detailed data to highly detailed data. 

The depth estimate technique we provide here, to our knowledge, is the only approach that can function on devices with resource constraints, e.g., Raspberry Pi, with accuracy on par with the state-of-art models and outperforms all models in both response time and battery utilization.

Our method does not depend on any underlying model or the type of ground truth (depth map). There is only one criterion for using our curriculum during training, which is the ground truth sparsity. To our knowledge, designing a curriculum for the KITTI dataset ~\cite{geiger2013vision} has not been studied.

There are growing pieces of evidence ~\cite{ma2019self,xiong2019foreground} that reconstructing the missing parts of sparse ground truth circumvents training challenges. The motivation behind reducing the ground truth sparsity originates from CNN depth estimations ~\cite{alhashim2018high} that demonstrate higher accuracy on datasets with dense ground truths than datasets with no dense ground truth. Although there are several promising approaches to reconstructing the missing pixels in a depth map ~\cite{rawassizadeh2019ghost, uhrig2017sparsity}, we have chosen a simple but very effective method that gives control over the extent to which the missing data should be reconstructed. 

To this end, we first sort 31 different versions of depth maps according to their sizes, which serve as a proxy for their complexity ~\cite{prechelt1998early} after a few max-pooling layers of various changes of kernel sizes. Later we describe our justification for having 31 different versions of the depth map.
%After arranging the syllabuses as a curriculum, we improve our approach by managing the complication of samples during the training phase.

Several promising research works ~\cite{jiang2015self,graves2017automated,hacohen2019power} show that training is faster and more accurate if samples in the training set are sorted in increasing order of complexity. This well-selected assortment of training samples, arranged based on prior knowledge, is called “curriculum” ~\cite{bengio2009curriculum}. Using a Curriculum learning improves training quantitatively and qualitatively ~\cite{pentina2015curriculum}. Besides, it improves the convergence rate and increases the likelihood of finding a better solution during optimization. 

Each training curriculum consists of a sequence of syllabuses that are either arranged automatically~\cite{graves2017automatic_curriculum} or sequenced-based training heuristics ~\cite{jiang2015self_paced_curriculum_learning}. We use the latter option during training according to the syllabuses in our curriculum; in later sections, we provide more details about our approach.

On the other hand, a great deal of work has been devoted to the initialization strategies of neural networks. In this work, we employ a model provided by Alhashim and Wonka ~\cite{alhashim2018high}, and we use ImageNet ~\cite{deng2009imagenet} pre-trained weights for the encoder and random weight initialization for the decoder, i.e. transfer learning. Note we use a dense-depth model for encoder-decoder, but our novelty is focused on the training process, which is completely different than state-of-the-art models. Other models receive data as batches, but we provide novel rules for preprocessing and implementing the algorithm instead of regular training.

The following lists our contributions to this work.
\begin{itemize}
    \item We employ curriculum learning that facilitates the training of models on vision data with sparse ground truth labels. Using sparse data to train a neural network has an issue with weight updates, i.e., most of the detailed derivation of the error backpropagation algorithm provides zero, and thus  weights are not updated accordingly. To overcome this issue, We implemented a novel training process. Our training process provides a strategy to evade sparsity. This approach enables the nonlinear function to become more complex, and as a result, the model learns better features.
    \item We propose a novel method that outperforms state-of-the-art methods in battery utilization and execution time. Besides, it operates with the smallest number of parameters in comparison to state-of-the-art methods. Withing experiments, We demonstrate that our method is the fastest and most battery-efficient depth estimation model. It could be implemented as an on-device model and enable devices with small and limited battery sizes to incorporate depth estimation. 
\end{itemize}

%add new works (papers with code)
\section{Related Work}
\label{sec:related_work}
 \subsection{Depth Estimation Algorithms} Practical depth estimation focuses mainly on prediction accuracy but also should consider the response time and computational complexity. In this section, we investigate depth estimation models from different aspects, including the approximate number of parameters. 
 
There are many research works for performing depth estimation on sparse data, but most of them are focused on the model architecture rather than the sparsity itself~\cite{DBLP:journals/corr/abs-1808-00769,fu2019monocular,DBLP:journals/corr/abs-1910-06727,DBLP:journals/corr/abs-1809-09061,DBLP:journals/corr/abs-2008-05158,8624583,uhrig2017sparsity}. 
Almost all models suffer from challenges created by their ground truth sparsity. Therefore, we did not follow a long-established path toward improving accuracy. Instead, we focus on data sparsity to find a solution for bypassing the challenges imposed by it.

\subsection{Task Formulation} Although depth estimation has mostly been formulated as a regression task ~\cite{eigen,alhashim2018high,bts,huynh2020guiding,hu2019revisiting,hao2018detail,xu2017multi,xu2018structured}, some approaches have recently cast this problem as a classification ~\cite{dorn,adabins}. We conjecture that it is possible to combine these two approaches in our proposed curriculum, leading to a loss function composed of two losses, each of which presents a different task formulation.

\subsection{Energy Efficient Machine Learning} Since resource efficiency is a salient characteristic of our approach. , in this section, we review a few resource-efficient machine learning approaches and methods, a.k.a, TinyML. Studies on-device machine learning is vast. 
Except for Transfer Learning~\cite{cass2019taking,cai2020tinytl}, which is an optimization in a new related task and allows increased performance and improved progress, curriculum learning develop a resource-efficient approach and optimized operations of the model.

Most studies in this direction employ a variety of convergent and optimization techniques such as byte and sub-byte integer quantization \cite{choi2018pact}. 
To overcome the severe limitations in terms of memory and battery, several approaches were proposed. Tools such as Google TensorFlow Lite  \footnote{https://www.tensorflow.org/lite} that these tools employ hardware-independent techniques such as compression and quantization after training ~\cite{capotondi2020cmix,ghamari2021quantization}.

Additionally, some approaches use customized algorithms for small devices, such as association rule mining ~\cite{raw2016scalable}, clustering ~\cite{raw2019indexing} or classification ~\cite{sefr}, but they are not specifically designed for depth estimation.

\subsection{Curriculum Learning}  
Surprisingly few research works have been devoted to the analysis of the impact of curriculum learning on depth estimation ~\cite{bengio2009curriculum}. We should emphasize that using a fixed preprocessing on ground truth data to reduce sparsity has been studied by ~\cite{uhrig2017sparsity}, but it is clear that it is not possible to come up with a curriculum using a one-shot technique. We show that carefully designing the sparse ground truth in a curriculum can significantly improve the model accuracy (see Table \ref{tab:syllabuses}).

\section{Methods}
\label{sec:proposed_method}
In this section, we present the proposed monocular depth estimation method that efficiently handles sparse depth data by introducing a novel training process. To study and handle sparsity, first, we focus on the imputation method for sparse-depth data, and then we employ curriculum design.

\subsection{Imputation of Sparse Data} 
Since our objective is to reduce the ground truth sparsity, we should look for an imputation method for pixels with invalid (less than $1e-3$) or missing depth, and this procedure define syllabuses. Algorithm \ref{alg:curriculum} presents our approach to generating each syllabus of the curriculum. Using this algorithm with different parameters leads to building ground truth data with different levels of complexity.
\begin{figure}[ht]
    \centering
    \includegraphics[width= 10 cm ]{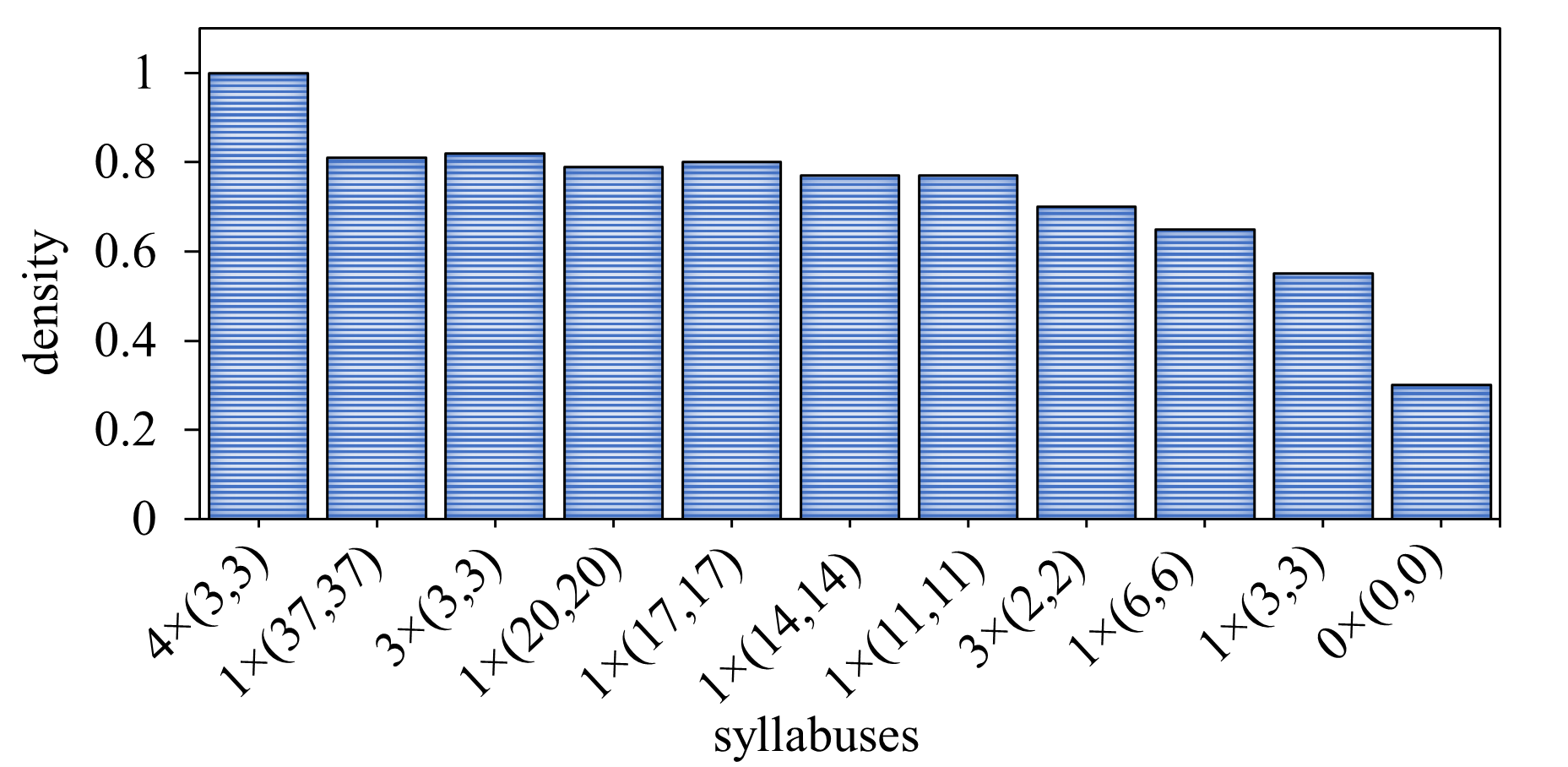}
    \caption{This histogram depicts data density (i.e., the ratio of valid pixels to all pixels in the ground truth) from a different iteration and max pooling kernel sizes. The histogram shows that as we cater to more complex syllabuses, the level of sparsity increases, reaching its maximum in the original ground truth depth map (column 0x(0,0)).}
    \label{fig:hist2}
\end{figure}
\paragraph{Apply Pooling and then Resizing.} Among diverse possible options for imputation of missing pixels, some of which are analyzed in section \ref{sec:ablation_study}, we empirically found that the 2D max pooling layer is the best approach for reducing the size of the depth map (lines 1 to 4 in Algorithm \ref{alg:curriculum}).

After 2048 ($256 \times 8$) max-pooling layers, we resize the output to its original size (lines 4 to 7 in Algorithm \ref{alg:curriculum}), which has more valid pixels (see Figure \ref{fig:hist2}) and also tweaking these parameters enables the model to interpolate between sparse and dense versions of the depth map. 

Besides, it is possible to progressively grow the size of our decoder based on different sizes of ground truth. Since it is not in the scope of this research we leave this strategy for future work. 

\begin{algorithm}[H]
  \caption{An Imputation Method($S, y, s$)}\label{alg:curriculum}
\textbf{Inputs:} training data $\train = \{(x,y)\}$ of size $n$ with sparse ground truth labels, and target size $s$ i.e., the model output size, a syllabus $S = (iterations,kernel\_size)$.  
  \begin{algorithmic}[1]
    \Procedure{Dilation}{$y, S, s$}
    \For{$i \in iterations$}
        \State $y = \textproc{MaxPool2D}(y, kernel\_size)$
    \EndFor
    \State $y = \textproc{Resize}(y, target\_size)$
    \State \textbf{return} $y$
    \EndProcedure
  \end{algorithmic}
\end{algorithm}

\paragraph{Syllabus Definition.} In each syllabus, we perform several iterations of the max pooling layer with a different kernel size to generate multi-resolution versions of ground truth (lines 1 to 4 in Algorithm \ref{alg:curriculum0}). Finally, based on our empirical analysis, we obtain 31 unique versions by removing duplicate versions that after resizing to the original size used to ground truth which we name the syllabus.

\begin{algorithm}[H]
  \caption{Different versions of depth maps($y, F, S$)}\label{alg:curriculum0}
\textbf{Inputs:} training data $\train = \{(x,y)\} $ of size $n$ with sparse ground truth labels, features $F = \{(iterations, kernel\_size)\:|\; m = min(y_{width},y_{height}), kernel\_size \in [2,m], iterations \in [1, log_2\left \lfloor m \right \rfloor]\}$ , $S$ is a set of unique syllabus.  
  \begin{algorithmic}[1]
    \Procedure{Preprocess}{$y, F, S$}
    \For{$i, k \in iterations, kernel\_size$}
        \State $y = \textproc{MaxPool2D}(y,(i, k))$
        \State $S.\textproc{Add}(y)$
    \EndFor
    \State \textbf{return} $S$
    \EndProcedure
  \end{algorithmic}
\end{algorithm}

\subsection{Curriculum Design} 
Our curriculum approach includes three components. First, sorting syllabuses with diverse levels of detail. Next, a policy specifies how to pass the syllabus to another syllabus. In the third part, by defining a patience term we enable the transfer from a syllabus to the next syllabus.

\begin{figure}[H]
    \centering
    \includegraphics[width=\textwidth]{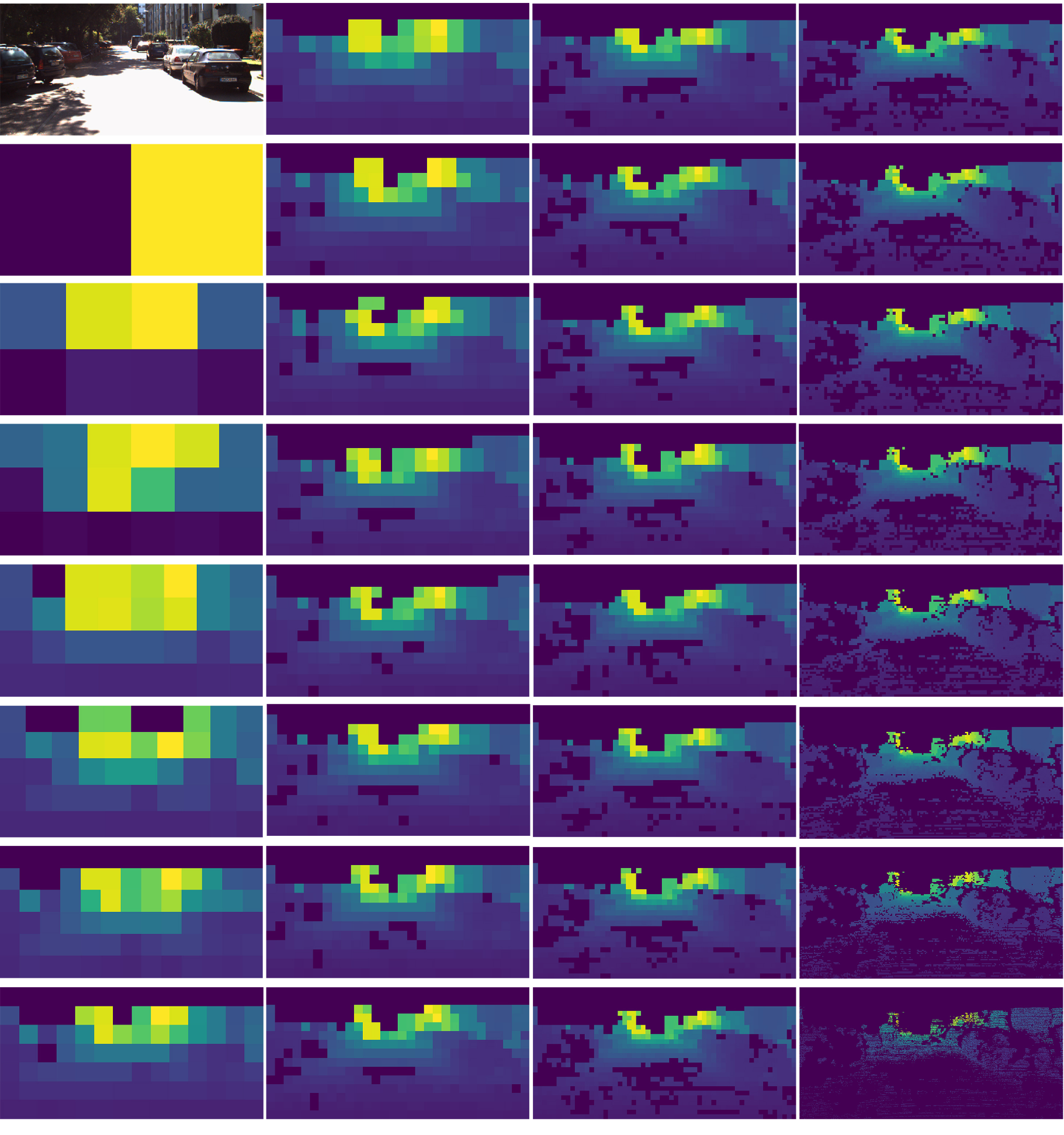}
    \caption{All possible choices of Tabe \ref{tab:syllabuses} to create a curriculum. The image in the bottom right corner is the original ground truth and the depth maps in each column, after the input (top left corner), are sorted according to their complexity.}
    \label{fig:curriculum}
\end{figure}

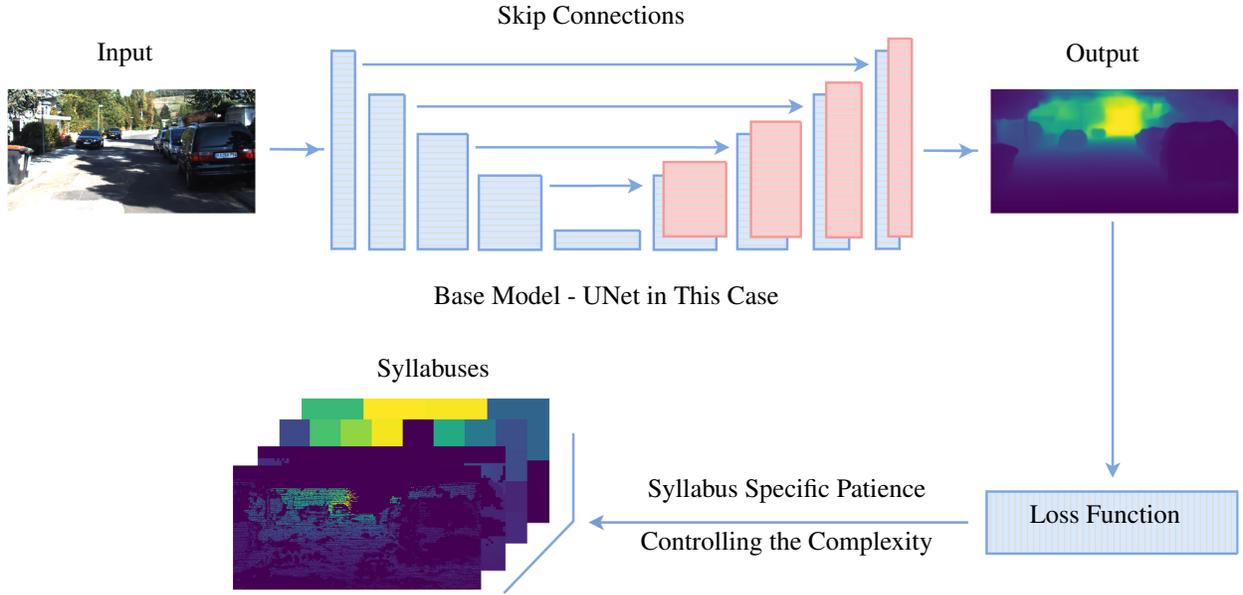
\begin{figure}[H]
    \begin{tikzpicture}
        \node[anchor=south west,inner sep=0] at (0,0) {\includegraphics[width=\textwidth]{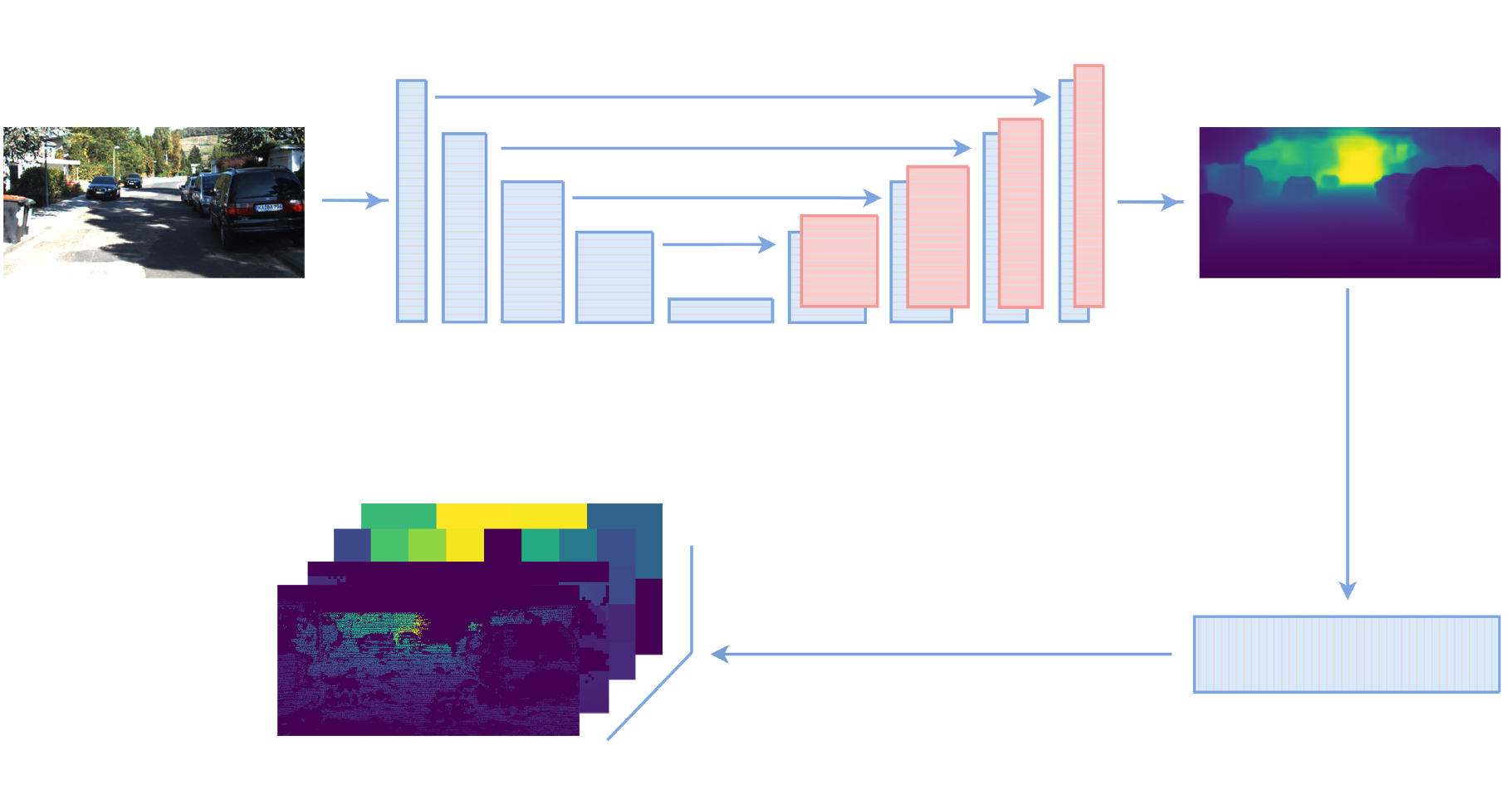}};
        \node at (1.6,7.7) {Input};
        \node at (7.8,8.2) {Skip Connections};
        \node at (8,4.5) {Base Model - UNet in This Case};
        \node at (14.6,7.7) {Output};
        \node at (14.63,1.6) {Loss Function};
        \node at (10.4,1.9) {Syllabus Specific Patience};
        \node at (10.4,1.2) {Controlling the Complexity};
        \node at (5.7,3.5) {Syllabuses};
    \end{tikzpicture}
    \caption{An overview of the training procedure with our proposed curriculum, in which, for every single input, there are multiple choices for the ground truth depth map. Each version of ground truth has a different level of sparsity, enabling our method to control the complexity of learning.}
    \label{fig:diagram}
\end{figure}

\paragraph{Sorting Syllabuses.} After defining the imputation method, we sort the syllabuses $S_i$ in an increasing order of complexity, i.e., based on their sparsity or the size of max pooling output. These syllabuses gradually address sparser versions of ground truth, such that the last syllabus leaves everything untouched and produces the original ground truth shown in Figure \ref{fig:curriculum}. Furthermore, the diagram of the whole process is presented in Figure \ref{fig:diagram}. The complete list of these syllabuses is available in Table \ref{tab:syllabuses}, but we only refer to a subset of these syllabuses, $\sA$, in the text, for our proposed curriculum. We defer the analysis of other possible choices to a later section \ref{sec:ablation_study}.

Afterward, we propose an algorithm (see Algorithm \ref{alg:curriculum2}) to automate the process of controlling the complexity of samples during the training. In this algorithm, we give a novel role to the early stopping patience.

\begin{figure}[H]    
    \begin{tikzpicture}
        \node[anchor=south west,inner sep=0] at (0,0) {\includegraphics[width=\textwidth]{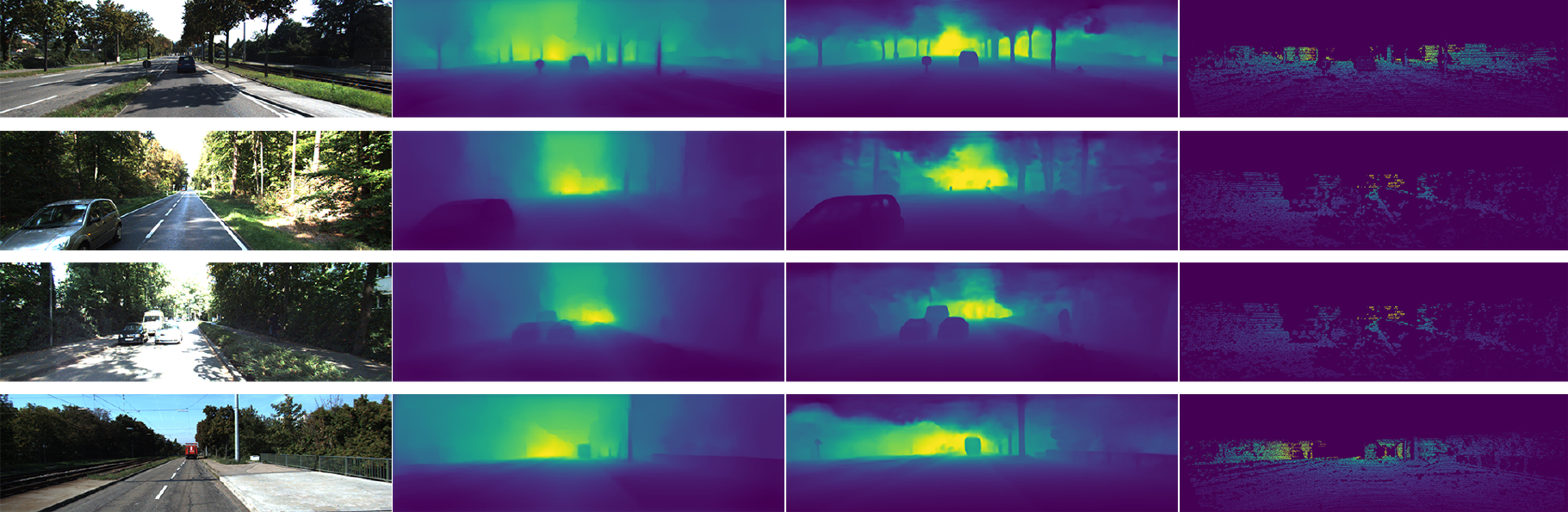}};
        \node at (14.4,-0.4) {Ground Truth};
        \node at (10.1,-0.4) {LightDepth};
        \node at (6,-0.4) {DenseNet};
        \node at (2,-0.4) {Input};
    \end{tikzpicture}
    \centering
    \caption{Visual comparison to demonstrate the improvement of our output over DenseNet ~\cite{alhashim2018high} on KITTI dataset. We have also included The left column presents the input and the right column presents sparse ground truth depth maps.}
    \label{fig:comparison}
\end{figure}
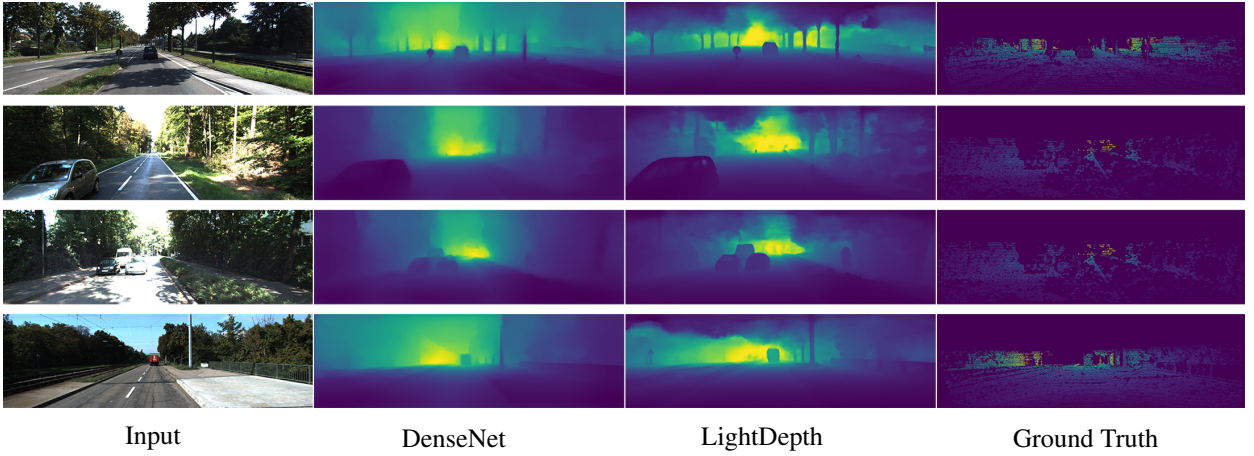

\paragraph{Passing Syllabuses.} Our model should stay in syllabus $S_i$ until it violates a condition for $P_i$ steps consecutively. The condition is described as follows: a decrease in the training loss should be higher than $\lambda$ percent of its current value. The parameter $\lambda$ is called the minimum decrease parameter. As we need syllabus-specific patience $P_i$ for each syllabus $S_i$ we denote a curriculum of size $k$ by a sequence of tuples $(S_i, P_i)_{i \in \sN_k}$, we present this process in lines 1 to 9 of Algorithm \ref{alg:curriculum2}.

\begin{algorithm}[H]
\caption{A Curriculum for Training on Sparse Labels($\lambda, (S_i)_{i\in \sN_k}, \train, s$)}\label{alg:curriculum2}
  \textbf{Inputs:} The minimum decrease parameter $\lambda \in [0,1]$, a sequence of syllabuses together with their corresponding patience $(S_i,P_i)_{i\in \sN_k}$ of size $k$ (or a curriculum of size $k$), a pretrained model $M$, training data $\train = \{(x_i,y_i)\}_{i \in \sN_n}$ of size $n$ with sparse ground truth labels, and target size $s$ i.e., the model output size.
  \begin{algorithmic}[1]
    \Procedure{TrainLoop}{$M, \lambda, (S_i)_{i\in \sN_k}, \train, s$}
    \For{$S, P \in (S_i, P_i)_{i\in \sN_k}$}
        \State $patience = 0$
        \For{$(x, y) \in \train$}
            \State $\hat{y} = \textproc{Dilation}(y,S, s)$ 
            \State $loss = \textproc{TrainStep}(M,x,\hat{y})$
            \State $train\_history.\textproc{Append}(loss)$
            \If{$train\_history[-1] > \lambda \cdot train\_history[-2] $}
                \State $patience = patience + 1$
                \If{$patience >= P$}
                    \State \textbf{break}
                \EndIf
            \EndIf
        \EndFor
    \EndFor
    \State \textbf{return} $M$
    \EndProcedure
  \end{algorithmic}
\end{algorithm}

\paragraph{Syllabus-specific Patience.} The syllabus-specific patience that we proposed is  similar to early stopping patience ~\cite{prechelt1998early}, but it has two major differences. First, it uses training loss instead of validation loss. Second, rather than breaking the training loop, it causes a shift to the next syllabus (line 9 in Algorithm \ref{alg:curriculum2}).

\section{Experiments and Results}
\label{sec:results_eval}
To present the efficiency of our method, we conducted experimental results on monocular depth prediction. After providing the implementation details of our approach, we describe the KITTI dataset and provide evaluation metrics on it. Then, we present an ablation study to discuss a detailed analysis of our approach and consider quantitative comparisons and some qualitative results with competitors.

\subsection{Implementation Details}

To demonstrate our curriculum, we trained a model for $106 \rm k$ steps called DenseDepth, which has a UNet ~\cite{ronneberger2015u} architecture with DenseNet196 ~\cite{huang2017densely} as the encoder backbone, while the whole model contains about $42.6 \rm M$ parameters. We modified data loaders available both in Pytorch and TensorFlow to use our curriculum. The batch size was set to $16$, and training was done on the Eigen split ~\cite{eigen} of the KITTI dataset ~\cite{geiger2013vision}, which has $23 \rm k$ training images and $697$ test images. Our method down-sampled Ground truth images to $256 \times 512$ during the training phase and the original sizes were maintained for the testing phase. Besides, we made some minor changes to the original DenseDepth ~\cite{alhashim2018high} implementation, including using Sigmoid activation in the output layer and training with a different cost function. We used ADAM ~\cite{kingma2014adam} optimizer as a cost function, with an initial learning rate of $1e-4$ and multiplicative weight decay of $0.9$ at every $23 \rm k$ step.

\begin{table*}[t]
\caption{Performance comparisons of the KITTI Eigen split dataset~\cite{eigen} for state-of-the-art depth estimation models. Bold texts show the best results.}
\label{tab:metrics}
    \begin{center}
    \begin{tabular}{lccccccc}
    ${\rm Method}$ & $\delta_{\rm 1} \uparrow$ & $\delta_{\rm 2} \uparrow$ & $\delta_{\rm 3} \uparrow$ &  ${\rm Abs \: Rel} \downarrow$ & ${\rm Sq \: Rel} \downarrow$ & ${\rm RMS}  \downarrow$ &  ${\rm RMS} \: log  \downarrow$\\
    \hline
    Saxena et al~\cite{saxena} & $0.601$ & $0.820$ & $0.926$ & $0.280$ & $3.012$ & $8.734$ & $0.361$\\
    Eigen et al.~\cite{eigen} & $0.702$ & $0.898$ & $0.967$ & $0.203$ & $1.548$ & $6.307$ & $0.282$\\
    Liu et al.~\cite{liu} & $0.680$ & $0.898$ & $0.697$ & $0.201$ & $1.548$ & $6.471$ & $0.273$\\
    Godard et al~\cite{godard} & $0.861$ & $0.949$ & $0.976$ & $0.114$ & $0.898$ & $4.935$ & $0.206$\\
    Kuznietsov et al.~\cite{semi} & $0.862$ & $0.960$ & $0.986$ & $0.113$ & $0.741$ & $4.621$ & $0.189$\\
    DenseDepth~\cite{alhashim2018high} & $0.886$ & $0.965$ & $0.986$ & $0.093$ & $0.589$ & $4.170$ & $0.171$\\
    Gan et al.~\cite{gan} & $0.890$ & $0.964$ & $0.985$ & $0.098$ & $0.666$ & $3.933$ & $0.173$\\
    DORN~\cite{dorn} & $0.932$ & $0.984$ & $0.994$ & $0.072$ & $0.307$ & $2.727$ & $0.120$\\
    Yin et al.~\cite{yin} & $0.938$ & $0.990$ & $0.998$ & $0.072$ & $-$ & $3.258$ & $0.117$\\
    BTS~\cite{bts} & $0.956$ & $0.993$ & $0.998$ & $0.064$ & $0.245$ & $2.756$ & $0.096$\\
    DPT-Hybrid~\cite{8dpt} & $0.959$ & $0.995 $  & $0.999$ & $0.062$ & $-$ & $2.573 $ & $0.091$\\
    LapDepth~\cite{7lap} & $0.962$ & $0.994 $  & $0.999$ & $0.059$ & $-$ & $2.446 $ & $0.093$\\
    AdaBins~\cite{adabins} & ${0.964}$ & ${0.995}$ & ${0.999}$ & ${0.058}$ & ${0.190}$ & ${2.360}$ & ${0.088}$\\
    GLPDepth~\cite{5glp} & $0.967$ & $0.996 $  & $0.999$ & $0.057$ &$-$ &  $2.297 $ & $0.086$\\
    DepthFormer~\cite{3depthformer} & $0.974$ & $0.997 $ & $0.999$ & $0.052$ & $0.158$ & $2.143 $ & $0.079$\\ 
    NeWCRFs~\cite{2new} & $0.975$  & $0.997$ & $0.999$ & $0.052$ & $0.155$ & $2.129$ & $0.079$\\
    BinsFormer~\cite{1binsformer} & $\mathbf{0.975}$ & $\mathbf{0.997}$ & $\mathbf{0.999}$ & $\mathbf{0.052}$ & $\mathbf{0.151}$ & $\mathbf{2.098}$ & $\mathbf{0.079}$\\
    LightDepth(Ours) & $0.940$ & $0.990$ & $0.997$ & $0.070$ & $0.294$ & $2.923$ & $0.111$\\
    \hline
    \end{tabular}
    \end{center}
\end{table*}

\begin{table*}[t]
\caption{Comparison with prior works in terms of the number of trainable parameters (Params), Gflops, Runtime, and Battery. Results on the KITTI Eigen split dataset~\cite{eigen} that were measured by the Raspberry Pi 4 device.  Parameters count in millions, Runtime in seconds, and Battery in Watt Seconds (WS). Bold texts show the best results.} 
\label{tab:params}
    \begin{center}
    \begin{tabular}{lccccccc}
    ${\rm Method}$ & ${\rm Params} \downarrow$ & ${\rm Gflops} \downarrow$ & ${\rm  Runtime(s) } \downarrow$ &  ${\rm  Battery}  \downarrow$ \\ 
    \hline
    DORN~\cite{dorn}  & $110.3 \rm M$ & $775.05$ & $60.6\pm 12.54$ & $363.06\pm75.24$\\ 
    BTS~\cite{bts}  & $112.8 \rm M$ & $299.32$ & $52.81\pm 0.34$ & $316.86\pm 2.04$ \\
    %~\citet{adabins}  & $78.2 \rm M$  & $300.66$ & $-$ & $- \rm M$ & $-$\\
    NeWCRFs~\cite{2new}  & $270.44 \rm M$  & $230.61$ & $69.23\pm 0.88$ &  $415.38\pm 6.88$\\
    DPT-Hybrid~\cite{8dpt} & $105.36\rm M$ & $103.76$ & $50.66\pm0.41$ & $303.96\pm2.46$ \\
    LapDepth~\cite{7lap} & $73.13 \rm M$  & $73.29$ & $55.35\pm 5.2$ & $232.1\pm 31.2$ \\
    GLPDepth~\cite{5glp} & $61.22 \rm M$  & $53.93$ & $45.22\pm 6.54$ & $271.32\pm 39.24$ \\
    LightDepth(Ours) & $\mathbf{42.6 M}$ &  $\mathbf{42.8}$  & $\mathbf{16.07\pm1.19}$ &  $\mathbf{ 96.42\pm7.14}$ \\
    \hline
    \end{tabular}
    \end{center}
\end{table*}

The depth estimation has been limited to an interval ranging from $1e-3$ to $80$ meters. Although the original implementation of DenseDepth only used random horizontal flips, we include more data augmentation methods, including random crop and random rotation during training.

\subsection{The KITTI Vision Benchmark Dataset.} To demonstrate the effect of using our curriculum, we used the KITTI Dataset, ~\cite{geiger2013vision} which consists of $27\rm k$ images from $61$ outdoor scenes. The images are of size $1241\times376$ and their corresponding ground truth depth maps, recorded by LiDAR sensors, are sparse. A few sample images of this dataset are provided in Figure \ref{fig:comparison}. Our statistical studies show that the majority of pixels in depth maps correspond to objects nearer than $50$ meters, and only around $25\%$ (see Figure \ref{fig:hist2}) of each depth map contains valid distances.

\subsection{Evaluation Result}

For evaluation, we apply the standard six metrics used by the previous works~\cite{eigen}:

Root mean squared error $(RMS):\sqrt{\dfrac{1}{N}\sum_{i}^n( d_{i} - \hat{d}_{i})^{2}}$, \\
Average relative error $(REL):\dfrac{1}{N}\sum_{i}^n \dfrac{\mid d_{i}-\hat{d}_{i}\mid}{d_{i}}$, \\ threshold accuracy $(\delta_i): \% of\; d_{i}\; s.t. \max(\dfrac{d_i}{\hat{d}_i}, \dfrac{\hat{d}_i}{d_{i}}) = \delta <  1.25, 1.25^2, 1.25 ^{3}$,\\ Squared Relative Difference$(sq.REL): \sqrt{\dfrac{1}{N}\sum_{i}^n \dfrac{\mid d_{i} - \hat{d}_{i}\mid}{d_{i}}}$; \\$(RMSlog): \sqrt{\dfrac{1}{N}\sum_{i}^n (log(d_{i}) - log(\hat{d}_{i}))^{2}}$.

where $d_{i}$ is a pixel in ground truth $\hat{d}_{i}$ is a pixel in the estimated depth image, and $n$ is the total number of pixels in the ground truth that are
available.

\subsection{Comparison to the state-of-the-art}
Our proposed method, LightDepth, achieves the same performance scores as state-of-the-art models while using fewer parameters. This is possible due to the use of our proposed curriculum, which results in faster convergence. In particular, we have fewer training steps (reducing 300k training steps ~\cite{alhashim2018high} to $106 \rm k$ training steps). Hence, our novel training procedure for sparse ground truth helps to use resources efficiently during both training and inference. As is shown in Table \ref{tab:params}, our model has outperformed all  state-of-the-art models by the number of parameters, Gflops,  execution time, and battery utilization.  We implement all models on Raspberry Pi v.4 \footnote{https://www.raspberrypi.com/products/raspberry-pi-4-model-b/}  and measure quantitative results (see Table \ref{tab:params}) under random tensors. Three state-of-the-art models ~\cite{adabins,1binsformer,3depthformer} failed to deploy on Raspberry Pi v.4, due to their high memory demand, and thus they have not been listed in Table \ref{tab:params}.

\begin{table}[H]
\caption{Complete List of Syllabuses} We can create a curriculum by selecting an arbitrary subset of the following list. Section \ref{sec:ablation_study} explains some possible choices for creating a curriculum, but the curriculum discussed in this paper is denoted by $\sA$. Moreover, the last syllabus generates the original depth map and should always be included in the curriculum to make training metrics comparable to testing metrics. The output sizes reported in this table, are before resizing depth maps back to their original size. The last number in the size of the depth map is the number of channels and $N$ stands for batch size.
\label{tab:syllabuses}
\begin{center}
\begin{tabular}{ccccc}
\multicolumn{1}{c}{\bf Index}&\multicolumn{1}{c}{\bf Iterations}  &\multicolumn{1}{c}{\bf Kernel Size}&\multicolumn{1}{c}{\bf Output Size}&\multicolumn{1}{c}{\bf Membership}\\
\hline
0          &8    &(2, 2)      &($N$, 1, 2, 1)& \\
1          &7    &(2, 2)      &($N$, 2, 4, 1)& \\
2          &4    &(3, 3)      &($N$, 3, 6, 1)& $\sA,\sB$ \\
3          &6    &(2, 2)      &($N$, 4, 8, 1)& \\
4          &2    &(7, 7)     &($N$, 5, 10, 1)& $\sB$\\
5          &1  &(37, 37)     &($N$, 6, 13, 1)& $\sA$\\
6          &2    &(6, 6)     &($N$, 7, 14, 1)& $\sB, \sC$\\
7          &5    &(2, 2)     &($N$, 8, 16, 1)& \\
8          &3    &(3, 3)     &($N$, 9, 18, 1)& $\sA,\sB$\\
9          &2    &(5, 5)    &($N$, 10, 20, 1)& $\sC$\\
10         &1  &(22, 22)    &($N$, 11, 23, 1)& $\sB$\\
11         &1  &(20, 20)    &($N$, 12, 25, 1)& $\sA$\\
12         &1  &(19, 19)    &($N$, 13, 26, 1)& $\sB,\sC$\\
13         &1  &(18, 18)    &($N$, 14, 28, 1)& \\
14         &1  &(17, 17)    &($N$, 15, 30, 1)& $\sA,\sB$\\
15         &4    &(2, 2)    &($N$, 16, 32, 1)& \\
16         &1  &(15, 15)    &($N$, 17, 34, 1)& $\sB,\sC$\\
17         &1  &(14, 14)    &($N$, 18, 36, 1)& $\sA$\\
18         &1  &(13, 13)    &($N$, 19, 39, 1)& $\sB$\\
19         &1  &(12, 12)    &($N$, 21, 42, 1)& $\sB$\\
20         &1  &(11, 11)    &($N$, 23, 46, 1)& $\sA,\sB,\sC$\\
21         &1  &(10, 10)    &($N$, 25, 51, 1)& \\
22         &2    &(3, 3)    &($N$, 28, 56, 1)& $\sB$\\
23         &3    &(2, 2)    &($N$, 32, 64, 1)& $\sA$\\
24         &1    &(7, 7)    &($N$, 36, 73, 1)& $\sB,\sC$\\
25         &1    &(6, 6)    &($N$, 42, 85, 1)& $\sA$\\
26         &1    &(5, 5)   &($N$, 51, 102, 1)& $\sB$\\
27         &2    &(2, 2)   &($N$, 64, 128, 1)& $\sC$\\
28         &1    &(3, 3)   &($N$, 85, 170, 1)& $\sA,\sB,\sC$\\
29         &1    &(2, 2)  &($N$, 128, 256, 1)& $\sC$\\
30         &-    & -  &($N$, 256, 512, 1)& $*$\\
\hline
\end{tabular}
\end{center}
\end{table}

Additionally, Figure \ref{fig:comparison} demonstrates some of the output samples that the edges are sharper and the outputs contain more details for the same model when incorporating our curriculum approach.

In this work, we do not devise a new encoder-decoder architecture, instead, we propose a novel efficient training process that outperforms the same encoder-decoder ~\cite{alhashim2018high}. With respect to the metrics of the base model(the blue row) used in our curriculum, there is $50\%$ improvement in ${\rm Sq \: Rel}$, $35\%$ in ${\rm RMS} \: log $, $29\%$ and $22\%$ in ${\rm RMS}$ and ${\rm Abs \: Rel}$.(see Table \ref{tab:metrics}).
%Table \ref{tab:params} complements the information provided in Table \ref{tab:metrics}, in which we compare the number of trainable parameters of each model together with its  Gflops, Runtime(s), Battery utilization, showing the efficiency of the model trained with our curriculum.

\subsection{Ablation Study}
\label{sec:ablation_study}

\paragraph{Curriculum Design}
Based on our experience, we identified that the choice of syllabuses plays a central role in the training procedure and the output quality. Therefore, it should be tuned like other hyperparameters. We have conducted a hyperparameter sensitivity analysis to find the optimal choice.

For example, curriculum $\sB$, with $16$ syllabuses, contains more than $\sA$, but we have found that increasing the number of syllabuses (more than that of $\sA$) does not improve metrics and decreases the convergence speed. While on the other hand, decreasing the number of syllabuses adversely impacts the best achievable presented metrics in Table \ref{tab:metrics}. Hence, we identify the optimal choice for the number of syllabuses as $11$.

While keeping the number of syllabuses unchanged, by introducing $\sC$, we further analyzed a different distribution of syllabuses in Table \ref{tab:syllabuses}, where the last (or more complex) items appear more often. We have found that skipping a few primaries (first syllabuses), and simple syllabuses not only reduces the convergence rate but sometimes makes it impossible to reach the best obtainable accuracy when they are included.
\paragraph{Base Model}
We have tested two models, including EfficientNet B5 ~\cite{tan2019efficientnet} and DenseNet121 ~\cite{huang2017densely} as the backbone of our Autoencoder to study the effect of having a different base model in our curriculum. We identify that given the huge increase in the number of trainable parameters, the performance gain is insignificant.

\paragraph{Imputation Method}
Although max pooling has been suggested as a proper way of imputation ~\cite{fu2019monocular}, we independently tested Gaussian filter and Mean pooling as possible layers for reconstructing the missing data. Mean pooling and Gaussian filter are not necessarily wind up with correct depth values. For example, the mean pooling layer includes the values of pixels that are zero, whereas we know that these pixels should be disregarded. 

\section{Conclusion}
\label{sec:conclusion}
Depth estimation algorithms have a wide range of applications. However, they are resource intensive, and not possible to deploy them on devices with limited computational capabilities.
Significant effort is devoted to devising more powerful depth estimation models by introducing extra complexity. However, in this research, we achieved competitive performance to the state-of-the-art models only by preprocessing the ground truth images through the formation of a curriculum to reduce sparsity at various levels.. designed process in the training phase helps to make an accurate prediction in an efficient way. Our approach outperforms all state-of-the-art models by their response time and battery utilization while maintaining accuracy. 
 
\bibliographystyle{abbrv}
\bibliography{arxiv/main}
\end{document}